\pdfoutput=1

\documentclass[11pt]{article}

\usepackage[final]{acl}
\usepackage{longtable}
\usepackage[utf8]{inputenc}
\usepackage{times}
\usepackage{mathrsfs}
\usepackage{latexsym}
\usepackage{geometry}
\usepackage{balance} 
\usepackage{amsmath}
\usepackage{amssymb} 
\usepackage{algorithm}  
\usepackage{algpseudocode}
\usepackage{subfigure}
\usepackage{subcaption}  
\usepackage{booktabs}  
\usepackage{multirow}  
\usepackage{tcolorbox} 
\usepackage{tabularx}
\tcbuselibrary{skins, breakable, theorems}
\usepackage{listings}
\usepackage{booktabs}
\usepackage{float}
\usepackage{url}
\usepackage{natbib}
\usepackage{xcolor}
\usepackage{color}
\usepackage{enumitem}
\usepackage{soul, color, xcolor}

\setlist[enumerate,1]{itemsep=0pt, topsep=0pt}

\usepackage[T1]{fontenc}
\usepackage[utf8]{inputenc}

\usepackage{microtype}

\usepackage{inconsolata}

\usepackage{graphicx}

\title{Probing the Difficulty Perception Mechanism of Large Language Models}

\author{
 \textbf{Sunbowen Lee\textsuperscript{1,2}},
 \textbf{Qingyu Yin\textsuperscript{3}},
 \textbf{Chak Tou Leong\textsuperscript{4}},
 \textbf{Jialiang Zhang\textsuperscript{1}},\\
 \textbf{Yicheng Gong\textsuperscript{2}},
 \textbf{Shiwen Ni\textsuperscript{5}},
 \textbf{Min Yang\textsuperscript{5}},
 \textbf{Xiaoyu Shen\textsuperscript{1}\thanks{Corresponding author}}
\\
\\
 \textsuperscript{1} Institute of Digital Twin, EIT ~
 \textsuperscript{2} Wuhan University of Science and Technology \\
 \textsuperscript{3} Zhejiang University  ~
 \textsuperscript{4} Hong Kong Polytechnic University\\
 \textsuperscript{5} Shenzhen Institutes of Advanced Technology, CAS\\
 \texttt{bw1863@outlook.com; xyshen@eitech.edu.cn}}

\begin{document}

\maketitle

\begin{abstract}
Large language models (LLMs) are increasingly deployed on complex reasoning tasks, yet little is known about their ability to internally evaluate problem difficulty, which is an essential capability for adaptive reasoning and efficient resource allocation. In this work, we investigate whether LLMs implicitly encode problem difficulty in their internal representations. Using a linear probe on the final-token representations of LLMs, we demonstrate that the difficulty level of math problems can be linearly modeled. We further locate the specific attention heads of the final Transformer layer: these attention heads have opposite activation patterns for simple and difficult problems, thus achieving perception of difficulty. Our ablation experiments prove the accuracy of the location. We uncover that there is a significant difference in entropy and difficulty perception at the token level, which provides practical support for using LLMs as automatic difficulty annotators, potentially substantially reducing reliance on costly human labeling in benchmark construction and curriculum learning. Overall, difficulty perception in LLMs is not only present but also structurally organized, offering new theoretical insights and practical directions for future research. Our code is available at \url{https://github.com/Aegis1863/Difficulty-Perception-of-LLMs}.
\end{abstract}

\section{Introduction}\label{introduction}
In the context of test time scaling, a large number of peer studies are focusing on how to enable models to adaptively control the output length based on the difficulty of the problem, which can avoid lengthy reasoning on simple problems \cite{openai2025gptoss, yang2025qwen3}. On the basis of DeepSeek-R1 \cite{deepseekr1}, people have mastered the method of using supervised fine-tuning combined with reinforcement learning (RL) for reasoning training. Therefore, the difficulty of adaptive token budget lies in how to label difficult and simple problems, as well as how to reduce the cost of this labeling. In this study, we focus on the mechanism of the model's perception of difficulty, as well as how to discover the simple and difficult problems that the model considers.

\begin{figure}[t]
    \centering
    \includegraphics[width=0.9\linewidth]{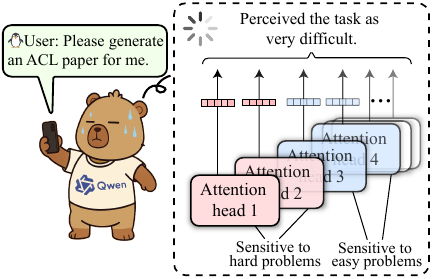}
    \caption{The LLM's perception of problem difficulty depends on the activation state of its specific attention heads.}
    \label{fig:first_page}
\end{figure}

\begin{figure*}[t]
    \centering
    \includegraphics[width=\linewidth]{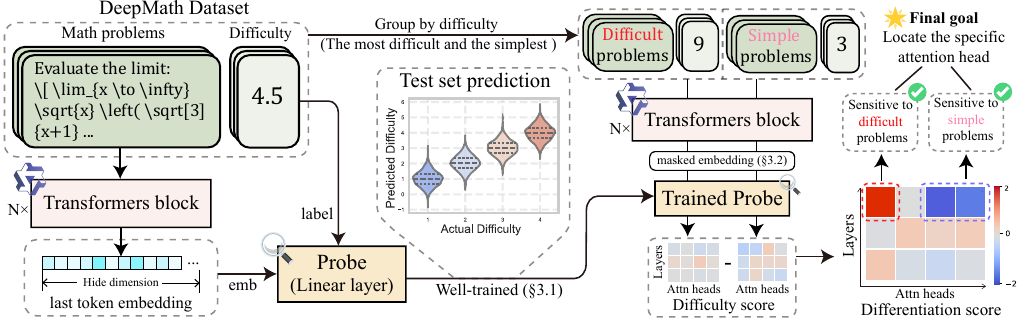}
    \caption{Probe training and attention heads pattern recognition. On the left we demonstrate how to train a difficulty probe based on last token embedding and corresponding difficulty labels. On the right, we show the process of using probe to identify attention head patterns, where the difficulty score of attention heads for difficult problems minus the difficulty score for simple problems yields a differentiation score, used to locate the attention heads most sensitive to difficulty.}
    \label{fig:framework}
\end{figure*}

Evaluating the difficulty of a question is inherently complex. First, a more reliable approach often involves human annotation, which, while generally more accurate, is both expensive and highly susceptible to subjective bias, and even model and human perceptions of difficulty may differ, making consistent and scalable difficulty evaluation challenging. Moreover, many existing methods attempt to bypass human labeling by using proxy metrics, such as the length of strong large language models' (LLMs) reasoning \cite{shi2025efficient}: the assumption being that longer responses indicate harder questions. However, this heuristic is frequently misleading. Powerful models like DeepSeek-R1 often produce redundant reasoning, which may reach the correct answer early but continue exploring alternative paths, inflating response length and distorting difficulty estimates. This issue is especially pronounced for open-ended questions (e.g., riddles or historical), where answer length bears little relation to actual cognitive difficulty. In general, the identification of problems through existing methods is not sufficiently reliable and lacks interpretability.

To address the lack of empirical evidence for difficulty perception, we, to the best of our knowledge, are the first to explore perceiving difficulty through the internal attention heads of LLMs. Leveraging DeepMath \cite{he2025deepmath103k}, a high-quality mathematical benchmark annotated with difficulty levels, we empirically demonstrate that well-trained LLMs can inherently encode the difficulty of mathematical problems. In particular, some models exhibit this perception directly through distinctive attention-head activation patterns. Internal perception in LLMs can be captured via a lightweight linear probe, while identifying a question’s difficulty merely requires tracking the activations of specific attention heads. Our mechanism interpretability experiments further reveal that LLMs linearly represent mathematical difficulty in a high-dimensional embedding space, and that such representations generalize beyond the training distribution.

We conduct token-level case studies on difficulty perception, which reveals several promising research directions. The analysis explores the relationship between perceived difficulty and entropy during inference. While previous work often assumes a correlation between question difficulty and entropy, the results show that internal difficulty perception of LLMs does not consistently align with entropy variation. Tokens identified as difficult also differ markedly from those with high entropy, suggesting that entropy-based difficulty estimation may not accurately capture the model’s internal evaluation.

In summary, the overall experimental procedure is illustrated in Figure \ref{fig:framework} and our contributions are as follows:
\begin{itemize}
    \item We prove that LLMs' perception of mathematical problems is high-dimensional linear and can accurately identify this direction.
    \item We precisely located the difficulty perception attention head of specific LLMs. Ablation experiments show that we can modify the perception by manipulating the output of the specific attention head.
    \item Our case study demonstrates inconsistencies in difficulty perception and entropy, and identifies tokens that are likely to cause significant changes in difficulty perception, providing promising insights for future research.
\end{itemize}

\section{Method}\label{sec:method}

Several peer studies have observed that, even before commencing reasoning, the embeddings generated by LLMs upon receiving a prompt can already reflect the intrinsic properties of that prompt, a phenomenon particularly pronounced in the domain of safety \cite{hu2024droj, lee2025xjailbreak, zheng2024prompt}. This observation has inspired our work on difficulty-perception modeling. 

\subsection{High-dimensional Linear Probe}\label{sec:method_probe}

\textbf{Low-dimensional representation is insufficient to clearly perceive difficulty.} Although \citet{zhu2025llmknows} has demonstrated that multimodal models such as Qwen2.5-VL \cite{bai2025qwen25vl} and InternLM-VL3 \cite{zhu2025internvl3} can significantly differentiate image-text pairs of different difficulty levels on low-dimensional representations at the last token embedding, this phenomenon cannot be observed intuitively in pure text scenarios. But we found that there is a difficulty-perception direction originally embedded in the high-dimensional linear space, which cannot be clearly observed in the low-dimensional space \cite{Lehengreasoning}, refer to Appendix \ref{appendix:lowdimensional}. However, the complex behavior of the LLMs may manifest as linear in high-dimensional embeddings. Based on the previous mechanism's explainability foundation \cite{hewitt, xu2024uncovering, arditi2024refusal, tighidet}, we believe that difficulty perception is high-dimensional linear.

\textbf{Training difficulty perception probe}. In order to investigate whether the problem embeddings encoded by the well-trained model can be consistent with the results annotated by humans, we employ a simple linear regression probe, a standard approach in representation probing \cite{tighidet}. Specifically, given a $d$-dimensional embedding vector $\mathbf{h}\in\mathbb{R}^d$ 
representing a question, the probe predicts a scalar difficulty score via a learnable linear transformation:
\begin{equation}
\hat{y}=\mathbf{w}^\top\mathbf{h}+b,
\end{equation}
where $\mathbf{w}\in\mathbb{R}^d$ and $b\in\mathbb{R}$ are trainable parameters, and $\hat{y}$ denotes the predicted difficulty. This probe is a standard linear regressor and is optimized by the minimum mean square error:
\begin{equation}
    \mathcal{L} = \frac{1}{N}\sum^N_{i=1}(y_i - \hat{y}_i)^2,
\end{equation}
where $y_i$ is ground-truth difficulty for the $i$-th question and $N$ is the number of training examples. We directly adopt a regression form for modeling without introducing an activation function, in order to obtain more granular difficulty perception results.

\subsection{Attention Head Pattern Recognition}\label{sec:method_attn_head}

Once the probe is well-trained, we can use it to identify specific attention head patterns.

To investigate the role of individual attention heads in encoding task difficulty, we propose a head-wise difficulty attribution framework based on selective attention ablation. Let $\mathbf{H} \in \mathbb{R}^{B \times L \times N \times d}$ denote the multi-head attention output, where $B$ is the batch size, $L$ the sequence length, $N$ the number of attention heads, and $d$ the head dimension. The final contextual representation is obtained via the output projection $\mathbf{W}_o \in \mathbb{R}^{(Nd) \times D}$, yielding 
\begin{equation}
\mathbf{Z} = \operatorname{Reshape}(\mathbf{H}) \mathbf{W}_o^\top \in \mathbb{R}^{B \times L \times D},
\end{equation}
with $D = Nd$ the model’s hidden dimension.

\begin{figure}[t]
    \centering
    \includegraphics[width=0.7\linewidth]{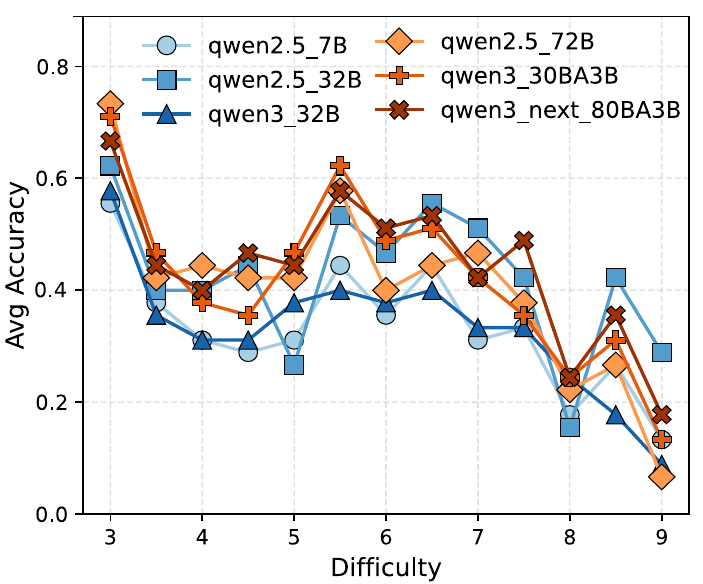}
    \caption{Accuracy rates of various models at different difficulty levels in DeepMath. As the difficulty increases, the accuracy rates of the models decrease, proving the rationality of manual labeling.}
    \label{fig:acc_deepmath}
\end{figure}

\begin{figure}
    \centering
    \includegraphics[width=0.7\linewidth]{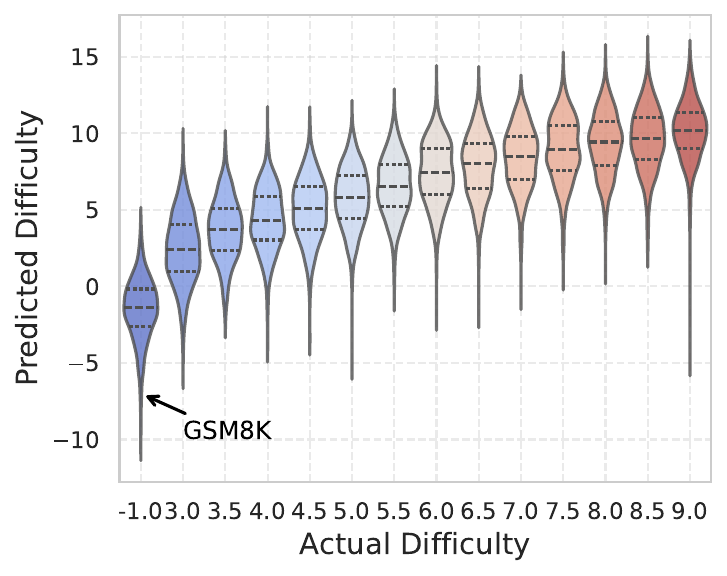}
    \caption{Probe test result. Among them, GSM8K is out of distribution data, showing lower prediction difficulty, consistent with expectations. The data on the right is the test set result of DeepMath. Since GSM8K does not have a real difficulty label, we symbolically mark it as -1 on the horizontal axis.}
    \label{fig:probe_test}
\end{figure}

We assume the availability of a pre-trained linear difficulty probe $\mathbf{v}_{\text{diff}} \in \mathbb{R}^D$, learned to predict scalar difficulty scores from the final-layer representation of the last input token. To isolate the contribution of the $i$-th attention head ($i \in \{1, \dots, N\}$), we construct an ablated representation $\mathbf{H}^{(i)}$ by zeroing out all heads except the $i$-th:
\begin{equation}
\begin{aligned}
\mathbf{H}^{(i)}_{b,\ell,j,:} = 
\begin{cases}
\mathbf{H}_{b,\ell,i,:}, & \text{if } j = i, \\
\mathbf{0}, & \text{otherwise},
\end{cases}
\\[4pt]
\forall\, b \in [B],\; \ell \in [L],\; j \in [N].
\end{aligned}
\end{equation}

The corresponding projected representation is $\mathbf{Z}^{(i)} = \operatorname{Reshape}(\mathbf{H}^{(i)}) \mathbf{W}_o^\top$. We extract the last token embedding $\mathbf{z}^{(i)}_b = \mathbf{Z}^{(i)}_{b, L-1, :} \in \mathbb{R}^D$ for each sample $b$, and compute its difficulty score as the normalized projection onto the difficulty direction:
\begin{equation}
s^{(i)}_b = \frac{ \langle \mathbf{z}^{(i)}_b, \mathbf{v}_{\text{diff}} \rangle }{ \| \mathbf{v}_{\text{diff}} \|_2 }.
\end{equation}

For a batch of samples sharing the same ground-truth difficulty level, we aggregate scores across the batch to obtain the mean head-wise difficulty attribution:
\begin{equation}
\bar{s}^{(i)} = \frac{1}{B} \sum_{b=1}^B s^{(i)}_b.
\end{equation}

By repeating this procedure over batches of varying difficulty (e.g., ``easy'' vs. ``hard'' instances), we derive a head-specific difficulty sensitivity profile. The discriminative power of head $i$ is quantified by the difference in its mean attribution between high- and low-difficulty cohorts:
\begin{equation}
\Delta^{(i)} = \bar{s}^{(i)}_{\text{hard}} - \bar{s}^{(i)}_{\text{easy}}.
\end{equation}

\begin{figure*}[t]
    \centering
    \includegraphics[width=0.9\linewidth]{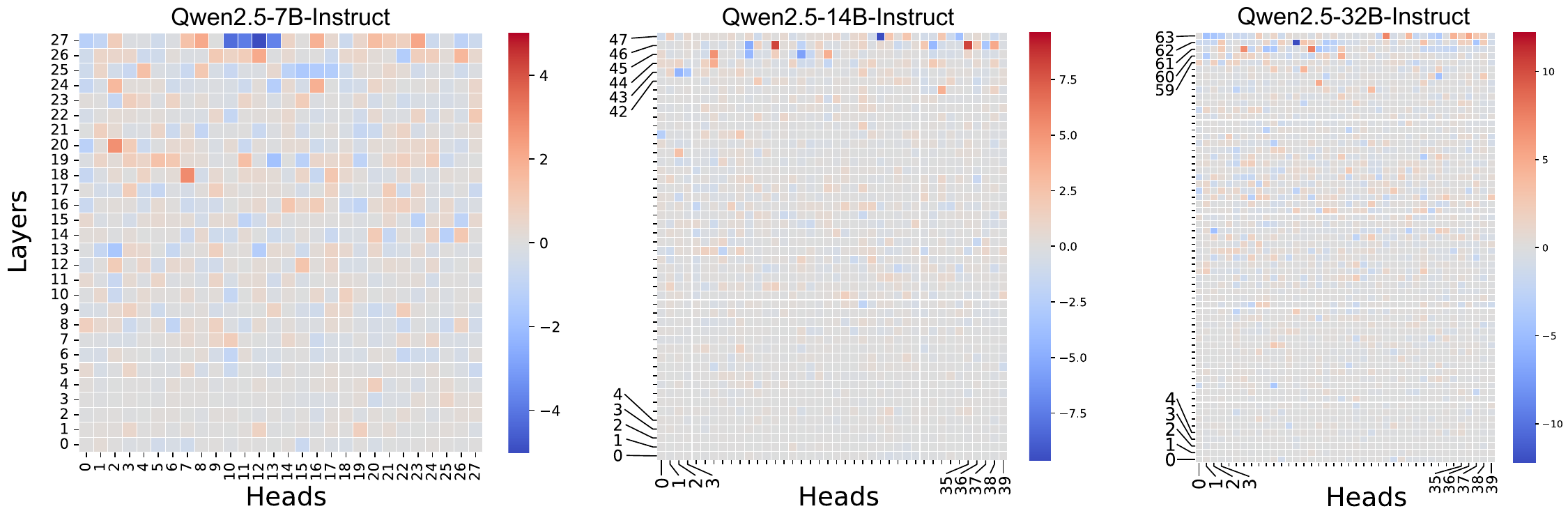}
    \caption{Different size model attention head pattern recognition results. We show the directions under significant difficulty differences (level 9 and 3). The blue attention head focuses on simple problems, while the red attention head corresponds to complex problem recognition.}
    \label{fig:qwen_attn}
\end{figure*}

This approach enables fine-grained analysis of how individual attention heads contribute to the model’s implicit difficulty perception, revealing specialized or redundant roles across the attention subspaces.

\section{Experiment}

\subsection{High-dimensional Linear Internal Difficulty Perception}

Utilizing the DeepMath \cite{he2025deepmath103k} dataset, which has meticulously annotated problem difficulty, we can train the linear probe mentioned in Subsection \ref{sec:method_probe}. We use various models \cite{qwen2025qwen25, yang2025qwen3} to test accuracy on this benchmark to verify the rationality of difficulty labels, and the experimental results are presented in Figure \ref{fig:acc_deepmath}, indicating general accuracy. Another reason for using this dataset is that it was released after the Qwen2.5 release, which minimizes the risk of data leakage to the greatest extent. Training loss and verification loss reference appendix Figure \ref{appendix:fig:qwen_probe_loss}, exhibits a normal convergence trend without overfitting tendencies.

We also present results on the DeepMath test set and include the GSM8K dataset \cite{cobbe2021gsm8k} as an out-of-distribution (OOD) benchmark. The DeepMath dataset originates from mathematical Olympiad problems and has been explicitly calibrated for difficulty, making it generally more challenging. In contrast, GSM8K consists of elementary school–level arithmetic problems. Although GSM8K lacks explicit difficulty annotations, its problem design suggests a substantially lower difficulty level compared to DeepMath. 

The train and test results in Figure \ref{fig:probe_test} and Appendix Figure \ref{appendix:fig:qwen_probe_loss} indicate that the probe can accurately categorize the mathematical problems in the DeepMath test set into clear difficulty levels, despite the presence of a small amount of long-tail data. At the same time, the data from GSM8K is as expected, located at a lower difficulty prediction level. We also trained a probe on llama3.1-8B-Instruct and still perform well, refer to Appendix \ref{appendix:llamaprobe}.

The above experiments on probe show that a trained LLM can linearly characterize the difficulty of the problem in high-dimensional space on last token embedding. In this case, we can obtain a difficulty-aware direction through the linear layer of the probe, and then guide the logit of the model in this direction. 

\subsection{Difficulty Perception Attention Head Localization}

Since the probe is effectively trained, we use the method in Subsection \ref{sec:method_attn_head} to identify difficulty-related attention head patterns (Figure \ref{fig:qwen_attn}). The color of each attention head indicates its preference for difficult (red) or easy (blue) problems, measured by the representation difference between difficulty levels 9 and 3.

Figure \ref{fig:qwen_attn} shows that difficulty perception mainly emerges in deeper transformer layers, while early layers exhibit little distinction across difficulty levels. The final layer contains the clearest patterns, suggesting it plays the dominant role in difficulty perception. For \textbf{Qwen2.5-7B-Instruct}, attention heads \textbf{10--13} specialize in easy problems, whereas heads \textbf{7, 8, 16, and 23} specialize in difficult problems. Although multiple heads jointly contribute in practice, individual head roles can be isolated by masking other heads and projecting the resulting representation onto the probe weights.

\begin{figure}[t]
    \centering
    \includegraphics[width=0.7\linewidth]{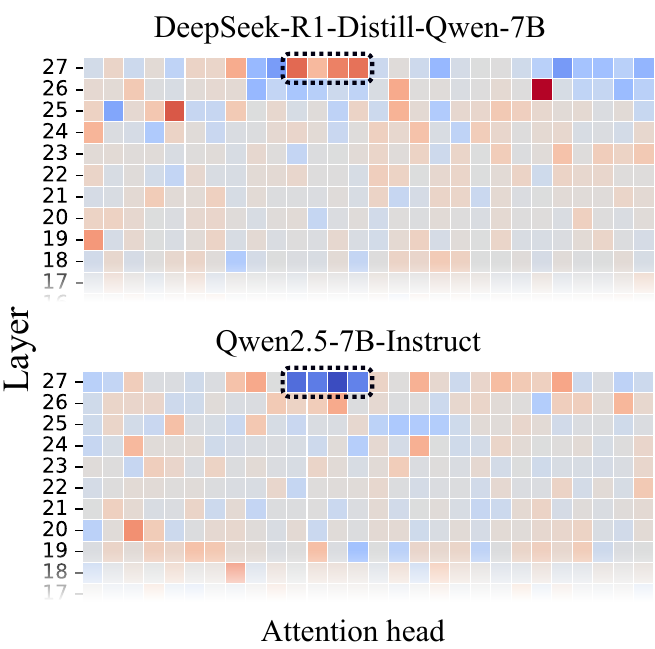}
    \caption{The attention head patterns of Qwen2.5-7B-Instruct and DeepSeek-R1-Distill-Qwen-7B. It is observed that the patterns of 4 attention heads are completely reversed, and the patterns of some other heads undergo partial changes.}
    \label{fig:different_qwen_distill}
\end{figure}

Figure \ref{fig:different_qwen_distill} compares Qwen2.5-7B-Instruct with DeepSeek-R1-Distill-Qwen-7B. After distillation, the primary perception heads (10--13) reverse their preference, and several other heads also change, indicating that training redistributes—but does not eliminate—the model's difficulty perception capability.

In contrast, Llama3.1-8B-Instruct exhibits no clear attention head pattern for difficulty perception (Appendix \ref{appendix:llamaattn}), suggesting that such mechanisms are model-dependent and may be influenced by pre-training.

In the next subsection, we validate these identified attention head patterns through ablation experiments.

\section{Attention Heads Analysis}

\subsection{Attention Heads Ablation}

To verify the functional specialization of attention heads revealed by our head-wise difficulty attribution analysis, we perform targeted ablation studies by intervening on heads associated with different difficulty levels. Using Qwen2.5-7B-Instruct, we isolate heads predominantly responsive to \textit{easy} problems and those most active on \textit{hard} ones. 

We perform two complementary interventions during inference:
\begin{itemize}
    \item \textbf{Difficulty increasing}: suppress the easy-mode heads by scaling their outputs by a factor of $0.1$, while enhancing the hard-mode heads by a factor of $2.0$;
    \item \textbf{Difficulty decreasing}: conversely, enhance the easy-mode heads ($\times 2.0$) and suppress the hard-mode heads ($\times 0.1$).
\end{itemize}

\begin{table}[t]
\centering
\scalebox{0.8}{\begin{tabular}{cccccc}
\toprule
\textbf{Real diff.} & \textbf{Original} & \textbf{Decrease} & \textbf{Increase} \\
\midrule
3.0 & 2.7 & 1.6 {\scriptsize\textcolor{green!60!black}{(-40.7\%)}} & 3.8 {\scriptsize\textcolor{red}{(+40.7\%)}} \\
4.0 & 3.5 & 2.1 {\scriptsize\textcolor{green!60!black}{(-40.0\%)}} & 4.8 {\scriptsize\textcolor{red}{(+37.1\%)}} \\
5.0 & 5.0 & 3.2 {\scriptsize\textcolor{green!60!black}{(-36.0\%)}} & 6.3 {\scriptsize\textcolor{red}{(+26.0\%)}} \\
6.0 & 5.6 & 3.8 {\scriptsize\textcolor{green!60!black}{(-32.1\%)}} & 6.8 {\scriptsize\textcolor{red}{(+21.4\%)}} \\
7.0 & 6.8 & 4.7 {\scriptsize\textcolor{green!60!black}{(-30.9\%)}} & 7.9 {\scriptsize\textcolor{red}{(+16.2\%)}} \\
8.0 & 7.6 & 5.4 {\scriptsize\textcolor{green!60!black}{(-28.9\%)}} & 8.5 {\scriptsize\textcolor{red}{(+11.8\%)}} \\
9.0 & 9.6 & 7.1 {\scriptsize\textcolor{green!60!black}{(-26.0\%)}} & 10.2 {\scriptsize\textcolor{red}{(+6.3\%)}} \\
\bottomrule
\end{tabular}}
\caption{Predicted difficulty under different adjustment conditions. Real diff. indicates the difficulty label from DeepMath's manual annotation. Original indicates the mean difficulty estimation output by the Probe when no intervention is applied; while Decrease represents the suppression of the difficulty perception of a specific attention head, and Increase indicates the corresponding enhancement.}
\label{tab:ablation}
\end{table}

These manipulations are implemented by head-wise multiplication of the corresponding head outputs in the multi-head attention tensor before the output projection:
\begin{equation}
    \mathbf{H}_{:, :, i, :} \leftarrow 
    \begin{cases}
        \alpha_{\text{reduce}} \cdot \mathbf{H}_{:, :, i, :}, & i \in \mathcal{S}_{\text{easy}}, \\
        \alpha_{\text{increase}} \cdot \mathbf{H}_{:, :, i, :}, & i \in \mathcal{S}_{\text{hard}}, \\
        \mathbf{H}_{:, :, i, :}, & \text{otherwise},
    \end{cases}
\end{equation}
where $\alpha_{\text{reduce}} = 0.1$, $\alpha_{\text{increase}} = 2.0$, $\mathcal{S}_{\text{easy}} = \{10,11,12,13\}$, and $\mathcal{S}_{\text{hard}} = \{7,8,16,23\}$. These attention heads are determined based on the experimental results shown in Figure \ref{fig:qwen_attn}.

As shown in Table~\ref{tab:ablation}, the difficulty increasing setting leads to a consistent increase in the model’s estimated difficulty scores across all inputs, effectively biasing the model toward perceiving problems as more challenging. Conversely, difficulty decrease yields lower difficulty estimates, aligning with the hypothesis that these head groups encode complementary signals for problem complexity. These results provide causal evidence that specific attention heads are functionally specialized for perceiving inputs of different difficulty levels.

Furthermore, we observe that if we manipulate the model to perceive simple questions as more challenging, the number of output tokens decreases, as shown in Table \ref{table:token_used_down}. Similarly, when the original model encounters difficult problems, the number of tokens in the output will also decrease. This shows a deficit of models like Qwen2.5-7B-Instruct, that is, when facing challenging problems, they tend to give up in advance rather than make more reasoning attempts. More test results are in Appendix \ref{appendix:RCTR}. 

\begin{table}[t]
\centering
\scalebox{0.8}{\begin{tabular}{ccc}
\toprule
 & \multicolumn{2}{c}{\textbf{Avg. token used}} \\ 
\cmidrule(lr){2-3}
\textbf{Real diff.} & \textbf{Original} & \textbf{Increase} \\
\midrule
3.0 & 809.23 & 692.23 {\scriptsize\textcolor{green!60!black}{(-14.5\%)}} \\
3.5 & 971.45 & 848.33 {\scriptsize\textcolor{green!60!black}{(-12.7\%)}} \\
4.0 & 1107.70 & 938.10 {\scriptsize\textcolor{green!60!black}{(-15.3\%)}} \\
4.5 & 1008.72 & 979.83 {\scriptsize\textcolor{green!60!black}{(-2.9\%)}} \\
5.0 & 1059.68 & 1015.08 {\scriptsize\textcolor{green!60!black}{(-4.2\%)}} \\
\bottomrule
\end{tabular}}
\caption{Average token usage under different difficulties. Original indicates the state without intervention. Increase refers to the state where humans make the model believe the problem is more challenging.}
\label{table:token_used_down}
\end{table}

\subsection{Intervention Validation}

\begin{table}[t]
\centering
\scalebox{0.65}{
\begin{tabular}{llccc}
\toprule
Intervention & Evaluation & Qwen & DPSK & Llama \\
\midrule
Selected (Hard$\rightarrow$Easy) & Probe B & $-0.135$ & $-0.653$ & $-0.273$ \\
Selected (Hard$\rightarrow$Easy) & Probe C & $+0.023$ & $-0.105$ & $-0.100$ \\
Norm-matched (Hard$\rightarrow$Easy) & Probe B & $-0.158$ & $+0.537$ & $-0.056$ \\
Random (Hard$\rightarrow$Easy) & Probe B & $-0.370$ & $-0.073$ & $+0.010$ \\
\midrule
Selected (Easy$\rightarrow$Hard) & Probe B & $-0.045$ & $+0.379$ & $+0.003$ \\
Norm-matched (Easy$\rightarrow$Hard) & Probe B & $+0.064$ & $+0.833$ & $-0.030$ \\
Random (Easy$\rightarrow$Hard) & Probe B & $-0.139$ & $+0.039$ & $-0.082$ \\
\bottomrule
\end{tabular}
}
\caption{Intervention validation using held-out evaluation probes. Probe B is trained on a disjoint split from head selection, while Probe C is an independent easy-versus-hard classifier. ``Norm-matched'' denotes heads matched by activation norm, and ``Random'' denotes randomly selected heads averaged over five seeds. Values are prediction shifts relative to the no-intervention baseline.}
\label{tab:intervention_validation}
\end{table}

To verify that the intervention is not an artifact of probe-based head selection, we separate the pipeline into disjoint data splits for probe training, head attribution, and evaluation. The selected heads are evaluated on an unseen split using Probe~B, which is trained independently of head selection, together with Probe~C, an easy-versus-hard classifier from a different probe family. As shown in Table~\ref{tab:intervention_validation}, the selected-head intervention consistently changes the predicted difficulty in the expected direction on DPSK and Llama, while Probe~C exhibits the same trend with smaller magnitude. In contrast, both random and activation-norm-matched controls produce substantially weaker effects or even the opposite direction, indicating that the observed behavior cannot be explained by generic activation scaling or random head selection.

\subsection{Behavioral Diagnostics}

\begin{table}[t]
\centering
\scalebox{0.78}{
\begin{tabular}{llcc}
\toprule
Model & Comparison & $\Delta$ Tokens & $p$ \\
\midrule
Qwen & Hard$\rightarrow$Easy vs None & $-108$ & $1.4\times10^{-4}$ \\
Qwen & Hard$\rightarrow$Easy vs Random & $-90$ & $2.2\times10^{-3}$ \\
Qwen & Hard$\rightarrow$Easy vs Norm & $-89$ & $7.4\times10^{-4}$ \\
Qwen & Easy$\rightarrow$Hard vs Norm & $+36$ & $2.5\times10^{-3}$ \\
\midrule
DPSK & Easy$\rightarrow$Hard vs Norm & $-101$ & $1.1\times10^{-5}$ \\
DPSK & Hard$\rightarrow$Easy vs Norm & $-49$ & $5.9\times10^{-3}$ \\
\midrule
Llama & Easy$\rightarrow$Hard vs None & $+267$ & $1.8\times10^{-4}$ \\
Llama & Easy$\rightarrow$Hard vs Random & $+365$ & $4.9\times10^{-8}$ \\
Llama & Easy$\rightarrow$Hard vs Norm & $+289$ & $1.9\times10^{-4}$ \\
\bottomrule
\end{tabular}
}
\caption{
Behavioral effects of the selected-head intervention measured by output token count. $\Delta$ Tokens denotes the mean difference in generated output length between the selected-head intervention and the comparison condition. Statistical significance is evaluated using paired Wilcoxon signed-rank tests over three decoding seeds per example.
}
\label{tab:behavior_results}
\end{table}

We additionally examine whether steering the identified heads produces measurable changes in generation behavior. Table~\ref{tab:behavior_results} reports paired Wilcoxon signed-rank tests on the generated output token count, comparing the selected-head intervention against the no-intervention baseline as well as random and activation-norm-matched controls. Here, $\Delta$ Tokens denotes the mean change in output length relative to the comparison condition. Although the direction of the effect varies across models, the selected-head intervention consistently produces statistically significant differences from the control conditions, suggesting that the identified heads influence downstream decoding behavior rather than merely perturbing hidden activations.

\subsection{OOD test on MATH500}

We evaluate the DeepMath-trained probe on the full MATH500 (500 problems with level 1–5 labels). No re-training is performed; the probe trained on DeepMath difficulty (3–9 scale) is applied directly.

\begin{table}[t]
\centering
\scalebox{0.92}{
\begin{tabular}{lcccc}
\toprule
Model & Spr. & PAcc & EH-AUC & Subj. Spr.\\
\midrule
Qwen  & \textbf{0.467} & 0.698 & \textbf{0.775} & 0.492 \\
DPSK  & 0.246 & 0.601 & 0.639 & 0.220 \\
Llama & \textbf{0.410} & 0.673 & \textbf{0.753} & 0.430 \\
\bottomrule
\end{tabular}
}
\caption{Difficulty prediction performance on the 5-level MATH benchmark. Spr. denotes Spearman correlation, PAcc denotes pairwise accuracy, EH-AUC denotes the AUC for distinguishing Levels 1--2 from Levels 4--5, and Subj. Spr. is the mean Spearman correlation over the seven MATH subjects. Higher is better for all metrics.}
\label{tab:math_level_results}
\end{table}

\begin{table}[t]
\centering
\scalebox{0.9}{
\begin{tabular}{lccccc}
\toprule
Model & L1 & L2 & L3 & L4 & L5\\
\midrule
Qwen  & 2.33 & 3.02 & 3.53 & 3.84 & \textbf{4.56} \\
DPSK  & 3.26 & 3.29 & 3.41 & 3.47 & 3.85 \\
Llama & 3.23 & 3.47 & 3.81 & 3.95 & 4.40 \\
\bottomrule
\end{tabular}
}
\caption{Average predicted difficulty score for each annotated difficulty level on the external 5-level MATH benchmark. L1--L5 correspond to the five annotated difficulty levels (sample counts: 43, 90, 105, 128, and 134, respectively).}
\label{tab:math_level_scores}
\end{table}

We further evaluate the latent probe on an external MATH benchmark with five annotated difficulty levels. As shown in Table~\ref{tab:math_level_results}, all models exhibit positive rank correlation with the ground-truth levels, with Qwen achieving the strongest performance, followed by Llama and DPSK. The average predicted scores increase monotonically across all five levels for every model (Table~\ref{tab:math_level_scores}), indicating that the probe preserves a consistent global ordering of problem difficulty. The per-subject Spearman correlation further shows that this ordering generalizes across the seven MATH subjects, while the weaker performance of DPSK suggests that its reasoning-distillation training compresses difficulty-related variation in the hidden representations.

\section{Case study}

Following the release of DeepSeek-R1, a large number of research efforts have extended on the basis of GRPO \cite{yu2025dapo, zheng2025group, wang20258020rule}. Many of them are hardworking attempts to implement an adaptive token budget based on difficulty, that is, to encourage model reasoning on difficult problems and suppress model overthinking on simple problems.

In this section, we discuss the law of difficulty perception during reasoning, as well as the relationship between token-level difficulty perception and entropy. This helps us better understand how the model considers the difficulty of problems during reasoning.

\subsection{Perception During Inference}

\begin{figure}[t]
    \centering
    \includegraphics[width=0.8\linewidth]{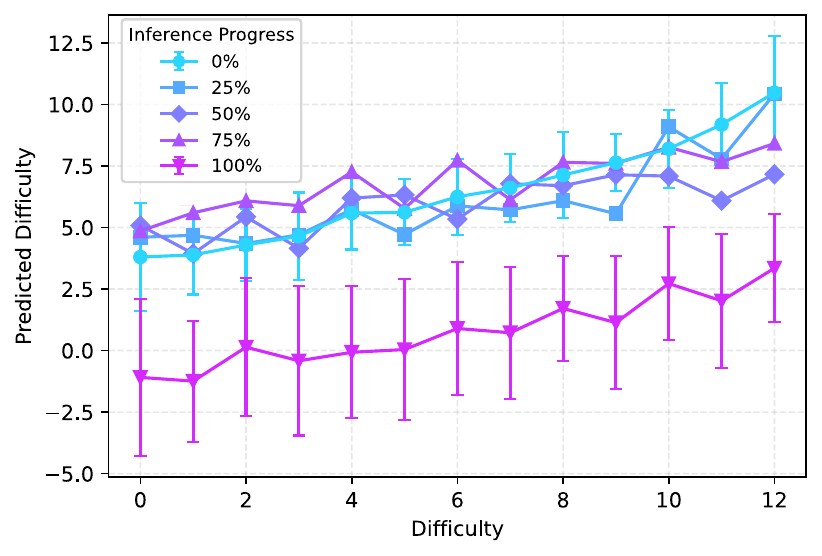}
    \caption{Utilize Probe for difficulty prediction during the inference process.}
    \label{fig:prober_infer}
\end{figure}

We first collect the complete responses of Qwen2.5-7B-Instruct for samples of various difficulty math questions. Then, we retain the parts of the responses according to their length percentage as 0\%, 25\%, 50\%, 75\%, and 100\%.

Referring to the experimental results in Figure \ref{fig:prober_infer}, we demonstrated the availability and predictive patterns of the probe at various truncations during inference. It can be observed that the estimated difficulty by the probe does not change significantly during the inference process and maintained the original trend. However, at the end of inference (100\%), its predicted values are significantly lower than those before, which is logical, as the problem has been solved at this point, and the model is not facing great difficulties. 

\subsection{Perceptual Difficulty and Entropy}

Entropy, as an indicator to measure the determinism of the model output, is widely used to refer to the difficulty of the question, which is usually true at the sentence level, but we find that it is not necessarily accurate at the token level.

\begin{figure}[t]
    \centering
    \includegraphics[width=0.85\linewidth]{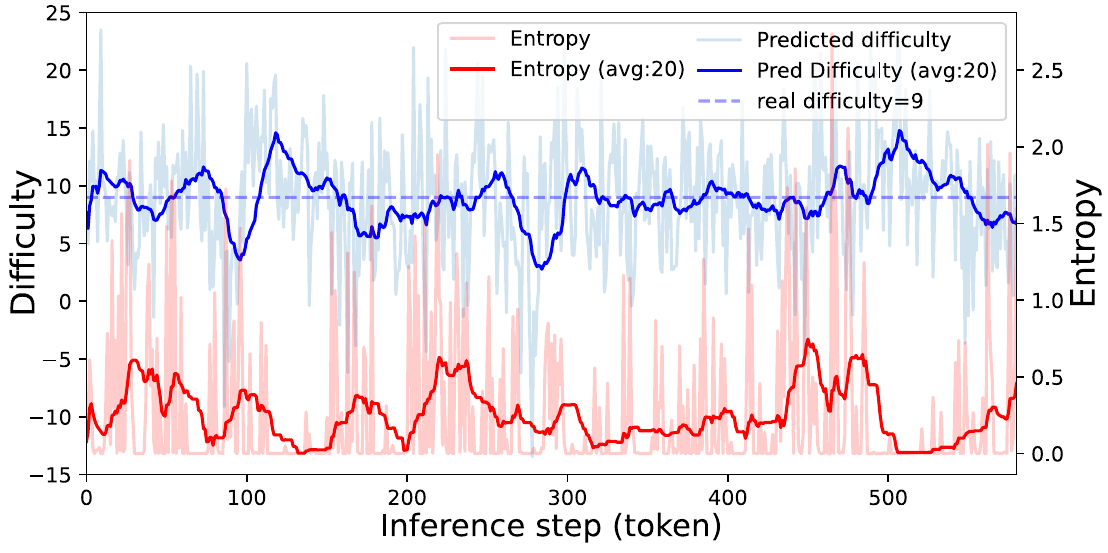}
    \caption{In the inference of a difficult math question (9.0), the change of entropy and difficulty perception.}
    \label{fig:entrpy_diff_1}
\end{figure}

In Figure \ref{fig:entrpy_diff_1}, it is observed that the difficulty perception curve varies around the ground-truth difficulty of 9.0, while its variation characteristics are not entirely equal to the change in entropy, that is, sometimes their trends are in the same direction, and sometimes they are completely opposite. The difficulty perceived by the model are not solely determined by the entropy of the next-token probability distribution.

\begin{figure}[t]
    \centering
    \includegraphics[width=0.83\linewidth]{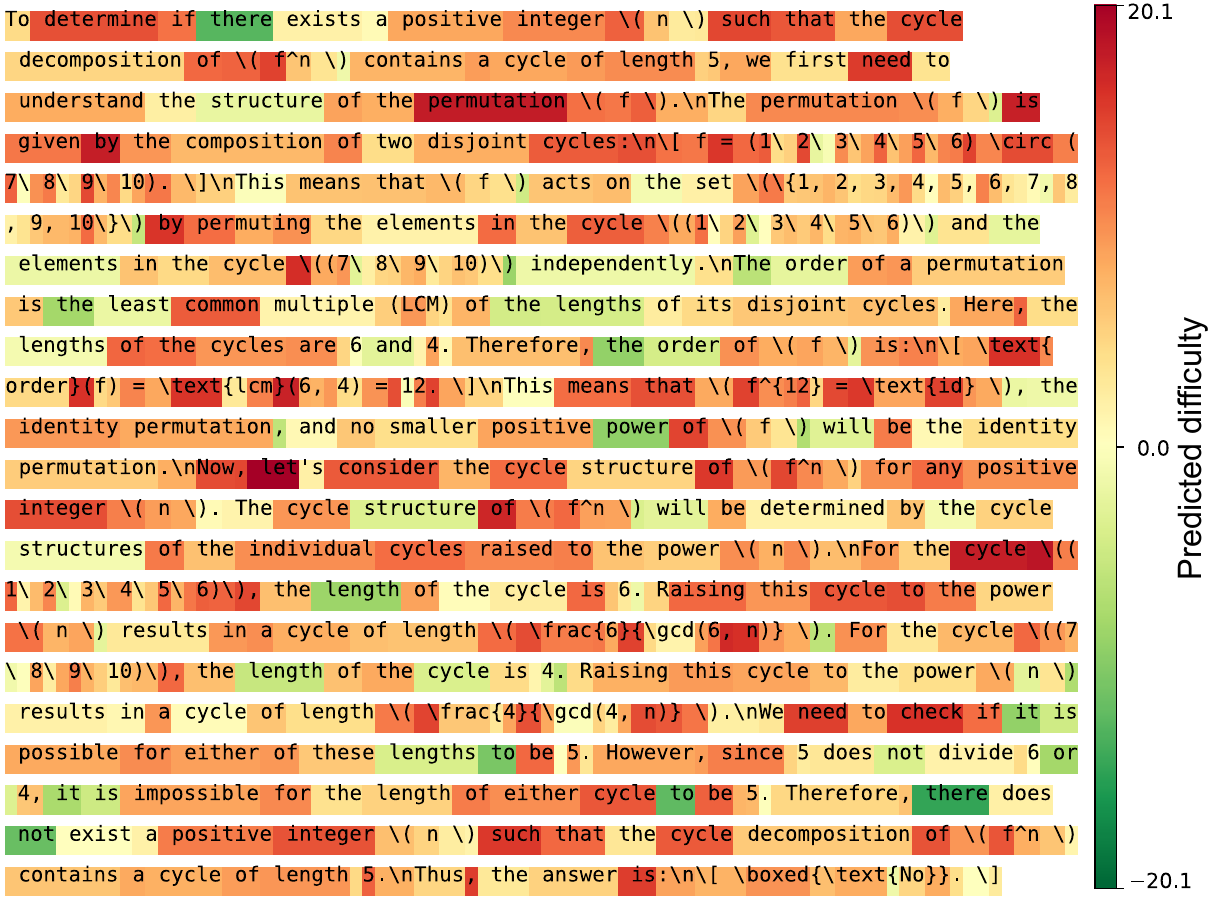}
    \caption{Token-level difficulty perception during inference.}
    \label{fig:difficulty_sentence}
\end{figure}

\begin{figure}[t]
    \centering
    \includegraphics[width=0.83\linewidth]{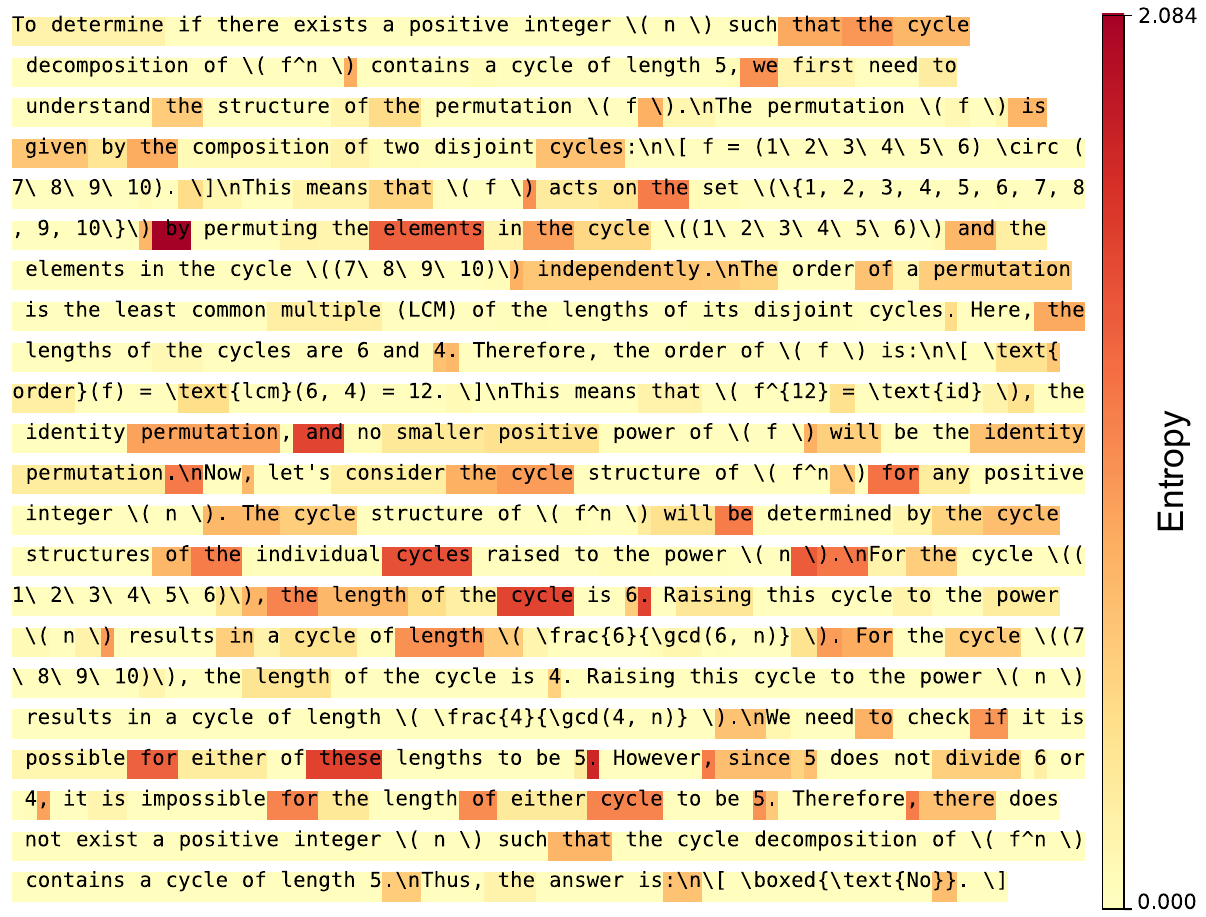}
    \caption{Token-level entropy during inference. To align with the color of Figure \ref{fig:difficulty_sentence}, we keep 0 as yellow.}
    \label{fig:entropy_sentence}
\end{figure}

In Figures \ref{fig:difficulty_sentence} and \ref{fig:entropy_sentence}, we demonstrate an experiment of token-level difficulty perception and entropy during inference. The closer to red indicates that the model currently considers the question difficulty to be greater (Figure \ref{fig:difficulty_sentence}), or the uncertainty of predicting the next token to be greater (Figure \ref{fig:entropy_sentence}). There is obviously a significant difference between the two, and in the perception based on difficulty, we found more granular results. An example is that our method shows greater difficulty when faced with numerical tokens, while in the results based on entropy, the entropy is mostly close to 0 when generating numbers, indicating a high degree of certainty. We believe that numbers are obviously worth paying attention to, as any error in numbers can have a huge impact on subsequent reasoning.

\subsection{Statistical Test of Entropy and Probe Difficulty}

\begin{table}[t]
\centering
\scalebox{0.85}{
\begin{tabular}{lccccc}
\toprule
Model & Probe & Entropy & Tokens & Probe$^\dagger$ & Entropy$^\ddagger$ \\
\midrule
Qwen & 0.846 & 0.599 & 0.002 & 0.100 & 0.372 \\
DPSK & 0.866 & 0.387 & 0.246 & 0.427 & 0.056 \\
Llama & 0.820 & 0.175 & 0.101 & 0.291 & -0.083 \\
\bottomrule
\end{tabular}
}
\caption{Spearman correlations with human-rated difficulty. Probe, Entropy and Tokens denote ordinary Spearman correlations. $^\dagger$Partial Spearman of the probe after controlling for entropy, prompt length and topic. $^\ddagger$Partial Spearman of entropy after controlling for the probe.}
\label{tab:entropy_partial}
\end{table}

We further investigate whether the latent probe simply reflects next-token uncertainty. Table~\ref{tab:entropy_partial} compares the probe with first-token entropy while controlling for prompt length and problem topic. The probe consistently retains positive partial correlation with difficulty after controlling for these factors, whereas the additional contribution of entropy becomes negligible for DPSK and Llama once the probe representation is known. An entropy-only regression baseline also performs substantially worse than the hidden-state probe, suggesting that the probe captures difficulty-related information beyond conventional uncertainty measures.

\section{Related Work}

Our work relates to research on difficulty perception in efficient reasoning, which can be broadly divided into \textit{external evaluation} and \textit{internal perception}.

\textbf{External evaluation.} Existing methods estimate problem difficulty using proxy signals. RL-based approaches dynamically allocate reasoning budgets or response lengths according to estimated difficulty \citep{luo2025o1pruner, shen2025dast, yang2025think}. Other proxies include the shortest correct solution \citep{yang2025thinking}, binary search over token budgets under a monotonicity assumption \citep{han2025token}, and dual reasoning modes that implicitly learn difficulty without explicit supervision \citep{wu2025arm, fang2025thinkless}. To reduce overthinking, \citet{huang2025hapo} truncated reasoning once a correct answer was produced.

\textbf{Internal perception.} Prior work suggests that LLMs implicitly encode difficulty-related information, but does not identify the corresponding attention head patterns. \citet{zhu2025llmknows} found that easy and hard problems cluster in representation space. Probing studies showed that safety-related concepts are linearly represented and can be manipulated through attention heads \citep{yin2025refusal, lee2025xjailbreak}. Likewise, activation steering demonstrates that internal representations can guide reasoning \citep{zbeeb2025reasoning, azizi2025activation}, while token confidence measures such as entropy and perplexity have been used to guide exploration \citep{ghasemabadi2025guided, wang20258020rule, cuietal2025stepwise}. Our work extends this line of research by explicitly identifying attention head patterns associated with difficulty perception.

\section{Conclusion}

In this study, we systematically investigated how LLMs perceive problem difficulty by constructing a probe based on mechanistic interpretability to identify high-dimensional difficulty representations. Trained on the human-annotated DeepMath dataset, the probe successfully revealed difficulty-related attention head patterns, and ablation experiments validated its effectiveness. We further showed that token-level difficulty perception differs substantially from entropy-based uncertainty, suggesting a complementary perspective on model reasoning. Additionally, our findings indicate several promising directions for future work: the clarity of difficulty-related attention patterns varies across models (e.g., Qwen2.5-7B-Instruct vs. Llama3.1-8B-Instruct), implying that pre-training and post-training quality may influence difficulty perception; these patterns can also evolve during training, as observed in DeepSeek-R1-Distill-Qwen-7B; and the relationship between token-level difficulty perception and entropy warrants further investigation.

\section*{Limitations}

This study still has some issues worth discussing, such as whether the method based on attention head difficulty perception can be used for reward allocation in the RL training process.

\section*{Ethics Statement and Usage Restrictions}

The attention head manipulation method used in this paper is solely for the purpose of studying the internal difficulty perception of LLMs. We do not support any malicious modification or misuse of the model. This study used a LLM for language polishing, and all writing has been manually proofread.


\bibliography{reference}

\appendix

\section{Low-dimensional Representation}\label{appendix:lowdimensional}

\begin{figure}[htbp]
    \centering
    \includegraphics[width=0.85\linewidth]{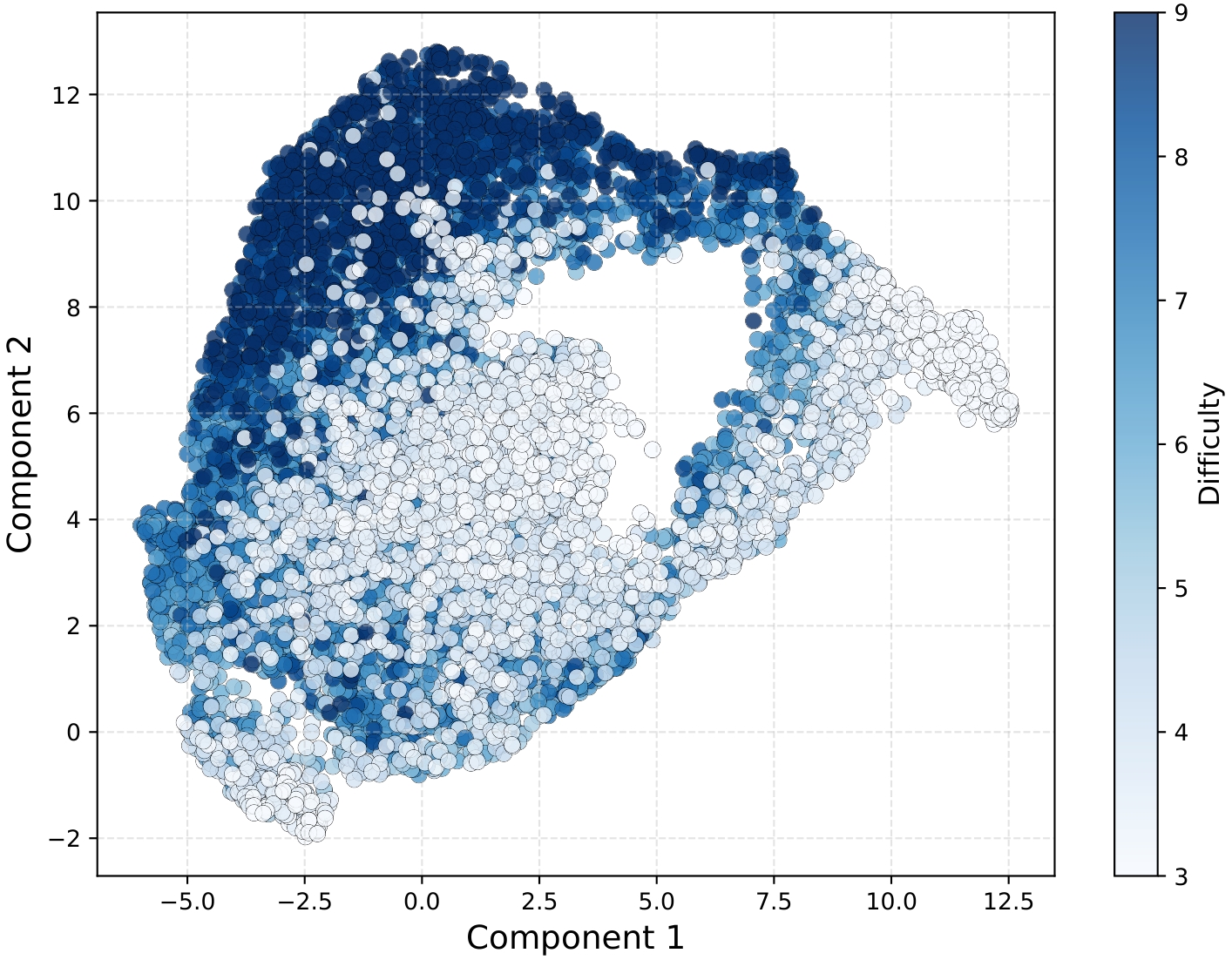}
    \caption{DeepSeek-R1-Distill-Qwen-7B's low-dimensional representation for DeepMath problems and difficulty is meticulously annotated by humans.}
    \label{appendix:fig:deepmath_dpsk_qwen_emb}
\end{figure}

\begin{figure}[htbp]
    \centering
    \includegraphics[width=0.7\linewidth]{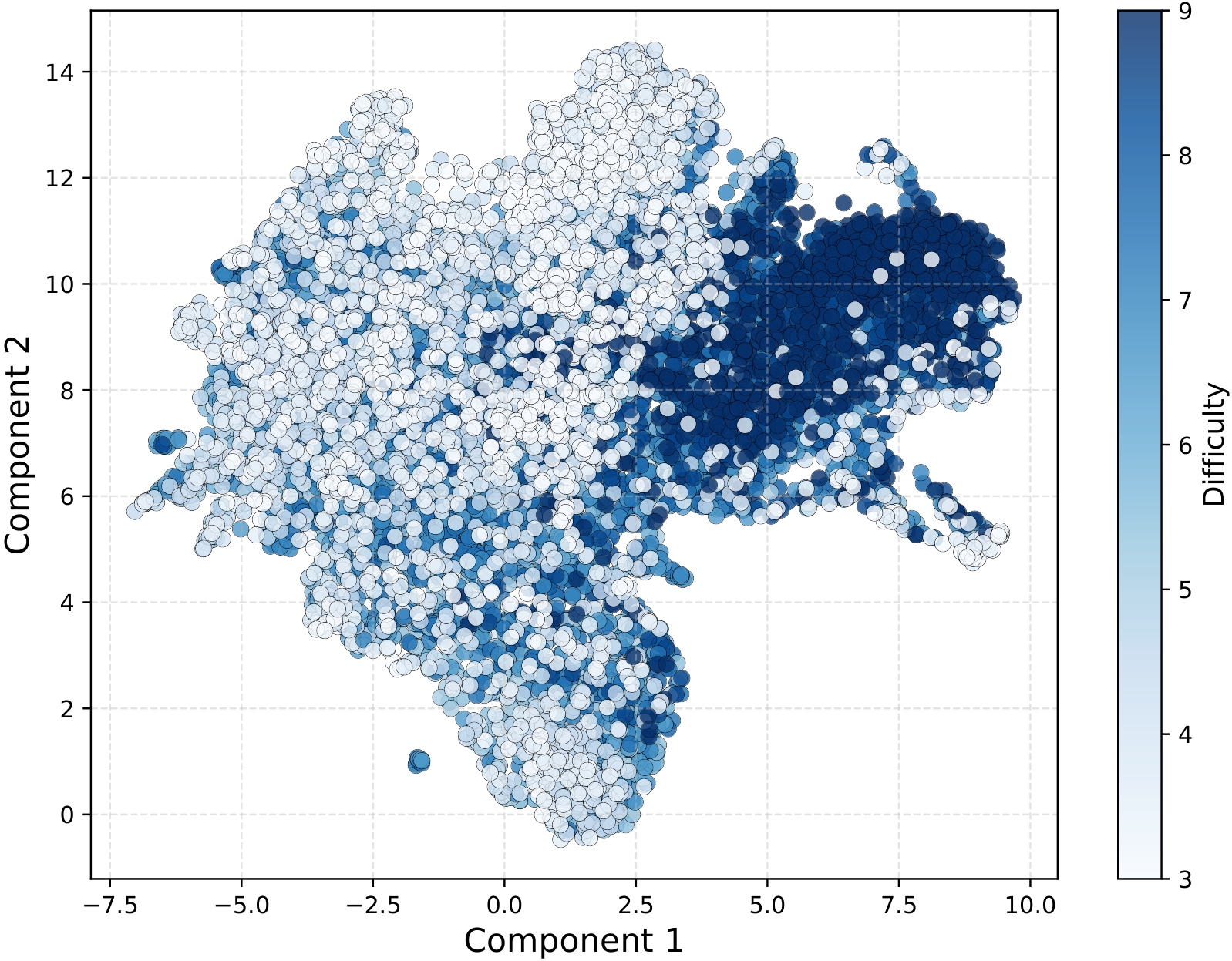}
    \caption{Qwen2.5-7B-Instruct's low-dimensional representation for DeepMath problems and difficulty is meticulously annotated by humans. Although difficult samples seem to come together, it is generally difficult to distinguish clearly.}
    \label{appendix:fig:deepmath_qwen_emb}
\end{figure}

\begin{figure}[htbp]
    \centering
    \includegraphics[width=0.85\linewidth]{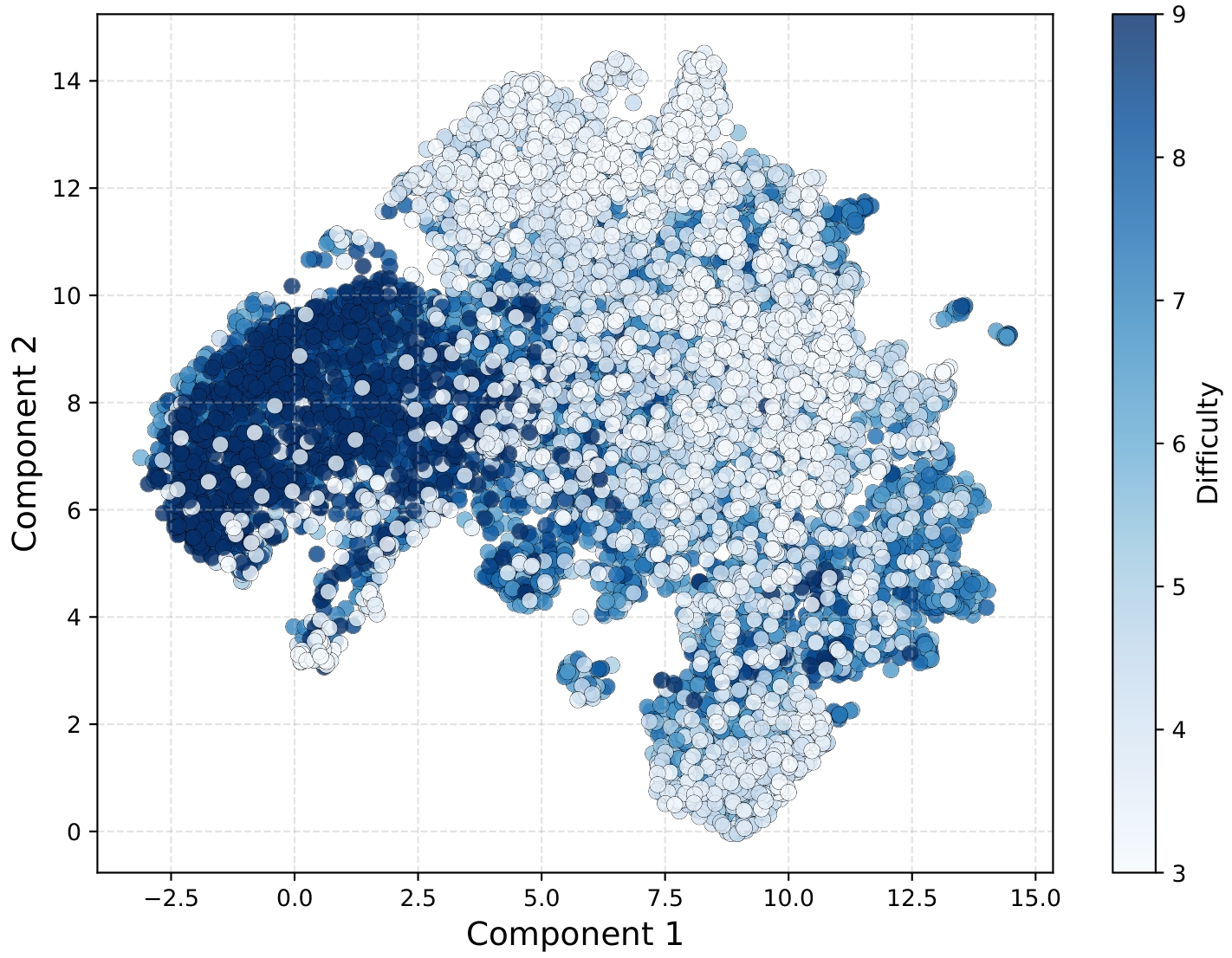}
    \caption{Tokenskip's low-dimensional representation for DeepMath problems and difficulty is meticulously annotated by humans.}
    \label{appendix:fig:deepmath_tokenskip_emb}
\end{figure}

We also demonstrated the low-dimensional representations of the DeepSeek-R1-Distill-Qwen-7B and TokenSkip \cite{xia2025tokenskip}, finding it difficult to clearly distinguish the spatial locations of different difficulty of questions. TokenSkip is a method of simplifying output by pruning tokens, allowing the model to respond to questions with more concise language. We tested TokenSkip to observe its impact on the model's difficulty perception. The experimental results show that there is little effect. Multiple low-dimensional representation image refer to Figure \ref{appendix:fig:deepmath_dpsk_qwen_emb}, \ref{appendix:fig:deepmath_qwen_emb} and \ref{appendix:fig:deepmath_tokenskip_emb}.

\section{Experiment Results of Llama3.1}\label{appendix:experiment}

We performed the same experiment on Llama3.1-8B-Instruct, which is shown in this section.

\subsection{Probe Training} \label{appendix:llamaprobe}

\begin{figure}[t]
    \centering
    \includegraphics[width=0.8\linewidth]{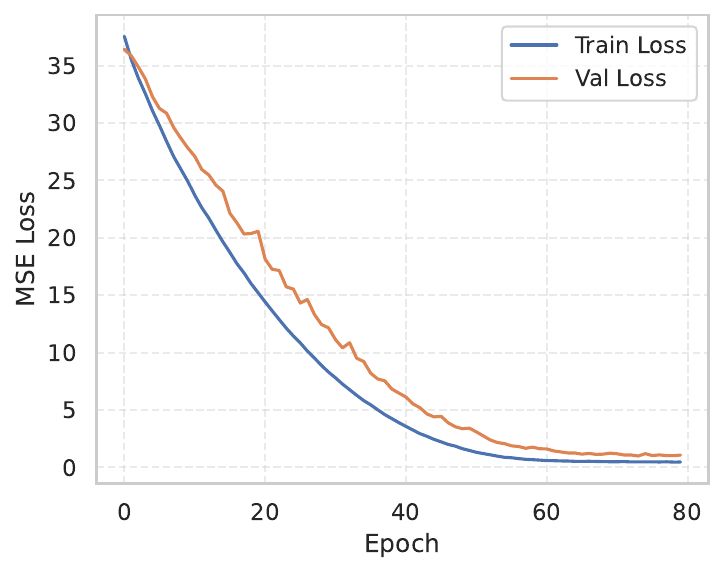}
    \caption{Probe training and validation loss. Training convergence is normal, no overfitting trend observed.}
    \label{appendix:fig:qwen_probe_loss}
\end{figure}

\begin{figure}[t]
    \centering
    \includegraphics[width=0.8\linewidth]{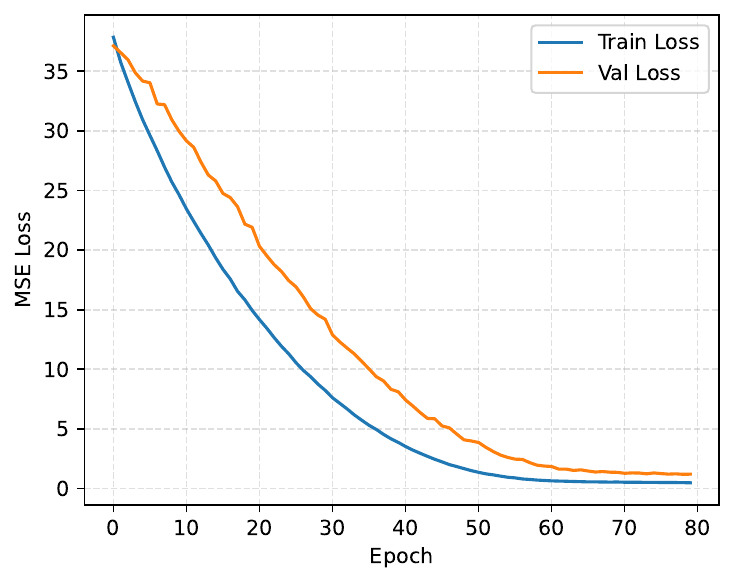}
    \caption{Llama3.1-8B-Instruct's probe training and validation loss.}
    \label{appendix:fig:llama_probe_loss}
\end{figure}

Qwen2.5-7B-Instruct and Llama3.1-8B-Instruct's probe training still has an good effect, refer to Figure \ref{appendix:fig:llama_probe_loss} and Figure \ref{appendix:fig:qwen_probe_loss}.

\subsection{Attention Head Pattern}\label{appendix:llamaattn}

\begin{figure}[t]
    \centering
    \includegraphics[width=0.8\linewidth]{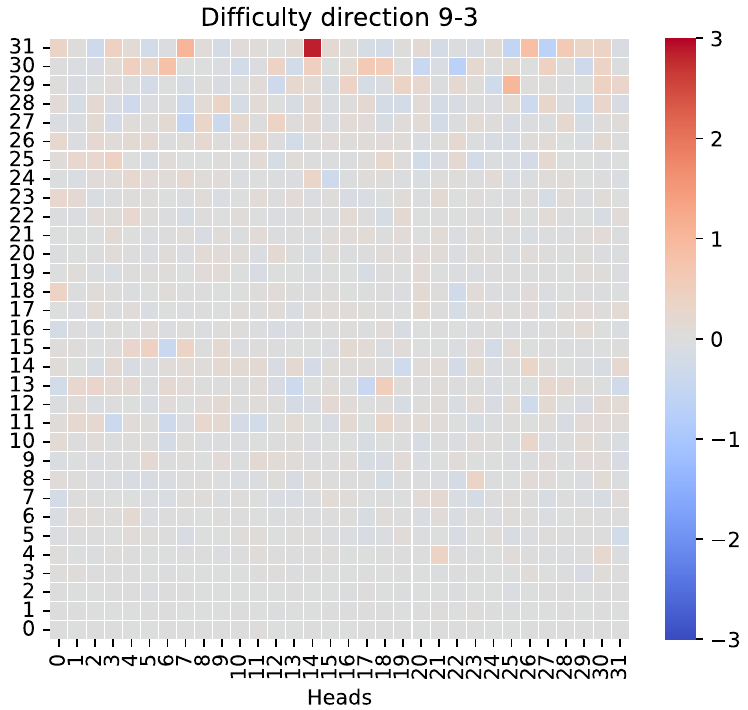}
    \caption{Llama3.1-8B-Instruct's attention head pattern recognition.}
    \label{appendix:fig:llama_attn}
\end{figure}

The Llama's attention head pattern is very insignificant, as referenced in Figure \ref{appendix:fig:llama_attn}. We can only observe that the 14th attention head in the last layer obviously serves to identify difficult problems. This is different from the Qwen series model, and we assume the reason is the pre-training quality.

\section{Reasoning Collapse Token Reduction} \label{appendix:RCTR}

Referring to Figure \ref{appendix:fig:llama_attn}, we found the reasoning collapse token reduction phenomenon. This means that when some models encounter very difficult problems, the number of tokens output dropped significantly, which is very similar to the findings of \citet{shojaee2025illusion}.

\begin{figure}[t]
    \centering
    \includegraphics[width=0.8\linewidth]{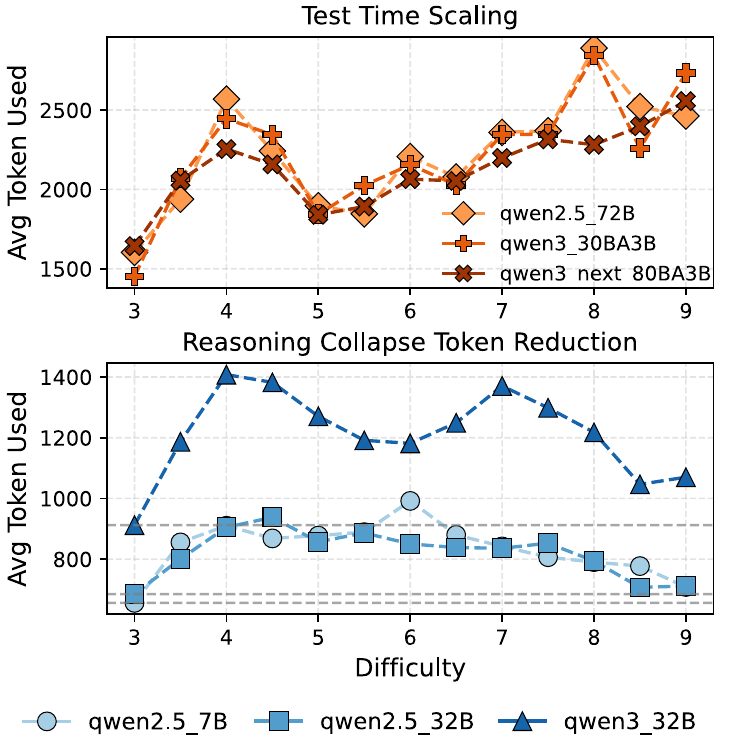}
    \caption{Reasoning Collapse Token Reduction. As the problems faced by the model become more difficult, the number of tokens it uses gradually increases. However, when some models reach a certain level of difficulty, the reasoning tokens they spend significantly decreased, while the other models continued to increase.}
    \label{appendix:token_used_difficulty}
\end{figure}

\end{document}